%% file: main.tex
\documentclass[10pt,twocolumn,letterpaper]{article}

\usepackage{cvpr}              

\usepackage{multirow}
\usepackage{rotating}
\usepackage{booktabs}
\usepackage{tabularx}
\usepackage{colortbl}
\usepackage{subcaption}
\usepackage[accsupp]{axessibility} 

\input{preamble}

\definecolor{cvprblue}{rgb}{0.21,0.49,0.74}
\usepackage[pagebackref,breaklinks,colorlinks,citecolor=cvprblue]{hyperref}

\usepackage[skins]{tcolorbox} 
\tcbuselibrary{breakable} 
\newtcolorbox[auto counter, number within=section, list type=subsubsection, list inside=toc]{sectionbox}[2][]{
colback=white!98!gray, colframe=black, 
colbacktitle=white!90!gray, coltitle=black, 
fonttitle=\bfseries,
title={#2}, 
list entry={Comment \thetcbcounter\quad}
}

\def\paperID{2046} 
\def\confName{CVPR}
\def\confYear{2024}

\title{OVER-NAV: Elevating Iterative Vision-and-Language Navigation with \\ Open-Vocabulary Detection and StructurEd Representation}

\author{Ganlong Zhao$^{1,2}$ \quad Guanbin Li$^{2,3}$\thanks{Corresponding authors are Guanbin Li and Yizhou Yu. This work was supported in part by the National Natural Science Foundation of China (NO.~62322608), in part by the CAAI-MindSpore Open Fund, developed on OpenI Community.} \quad Weikai Chen$^{4}$ \quad Yizhou Yu$^{1*}$\\
$^1${The University of Hong Kong} \quad $^2$Sun Yat-sen University\\ 
$^{3}$GuangDong Province Key Laboratory of Information Security Technology\\
$^4${Digital Content Technology Center, Tencent Games} \\
{\tt\small  zhaogl@connect.hku.hk, liguanbin@mail.sysu.edu.cn, chenwk891@gmail.com, yizhouy@acm.org}
}

\begin{document}
\maketitle
\input{sec/0_abstract}
\input{sec/1_introduction}

\input{sec/2_related}
\input{sec/3_method}

\input{sec/4_experiment}

\input{sec/5_ablation}
\input{sec/6_conclusion}
{
    \small
    \bibliographystyle{ieeenat_fullname}
    \bibliography{main}
}

\input{sec/X_suppl}

\end{document}

%% file: preamble.tex
\usepackage[dvipsnames]{xcolor}

\newcommand{\benchmark}{IR2R}

\newcommand{\thamt}{TourHAMT\xspace}

\newcommand{\methodnameshort}{OVER-NAV\xspace}
\newcommand{\methodnamelong}{OVER-NAV\xspace}
\newcommand{\keywordmap}{omnigraph\xspace}
\newcommand{\KeywordMap}{Omnigraph\xspace}

\newcolumntype{s}{>{\columncolor[gray]{.85}[.5\tabcolsep]}c}

\makeatletter
\DeclareRobustCommand\onedot{\futurelet\@let@token\@onedot}
\def\@onedot{\ifx\@let@token.\else.\null\fi\xspace}

\def\eg{\emph{e.g}\onedot} 
\def\ie{\emph{i.e}\onedot}

\makeatother

%% file: sec/0_abstract.tex
\begin{abstract}
Recent advances in Iterative Vision-and-Language Navigation~(IVLN) introduce a more meaningful and practical paradigm of VLN by maintaining the agent's memory across tours of scenes. Although the long-term memory aligns better with the persistent nature of the VLN task, it poses more challenges on how to  utilize the highly unstructured navigation memory with extremely sparse supervision. Towards this end, we propose OVER-NAV, which aims to go over and beyond the current arts of IVLN techniques. In particular, we propose to incorporate LLMs and open-vocabulary detectors to distill key information and establish correspondence between multi-modal signals. Such a mechanism introduces reliable cross-modal supervision and enables on-the-fly generalization to unseen scenes without the need of extra annotation and re-training. 
To fully exploit the interpreted navigation data, we further introduce a structured representation, coded Omnigraph, to effectively integrate multi-modal information along the tour. Accompanied with a novel omnigraph fusion mechanism, OVER-NAV is able to extract the most relevant knowledge from omnigraph for a more accurate navigating action. In addition, OVER-NAV seamlessly supports both discrete and continuous environments under a unified framework. We demonstrate the superiority of OVER-NAV in extensive experiments.

\end{abstract}

%% file: sec/1_introduction.tex
\section{Introduction}
\label{sec:intro}


{Vision-and-Language Navigation~(VLN)}~\cite{anderson2018vision} aims to build intelligent agents that can follow natural language instructions to navigate in the unseen environments. 
However, existing VLN benchmarks eliminate the agent's memory upon the start of every episode, failing to leverage the visual observations and the iteratively built maps collected by the physical robots. 
Recent work on {Iterative Vision-and-Language Navigation~(IVLN)}~\cite{krantz2023iterative} introduces a more meaningful and practical paradigm that orders the episodic tasks in VLN as tours and allows the agents to utilize memory to achieve better navigation performance. 


\begin{figure}[t]
  \centering
   \includegraphics[width=\linewidth]{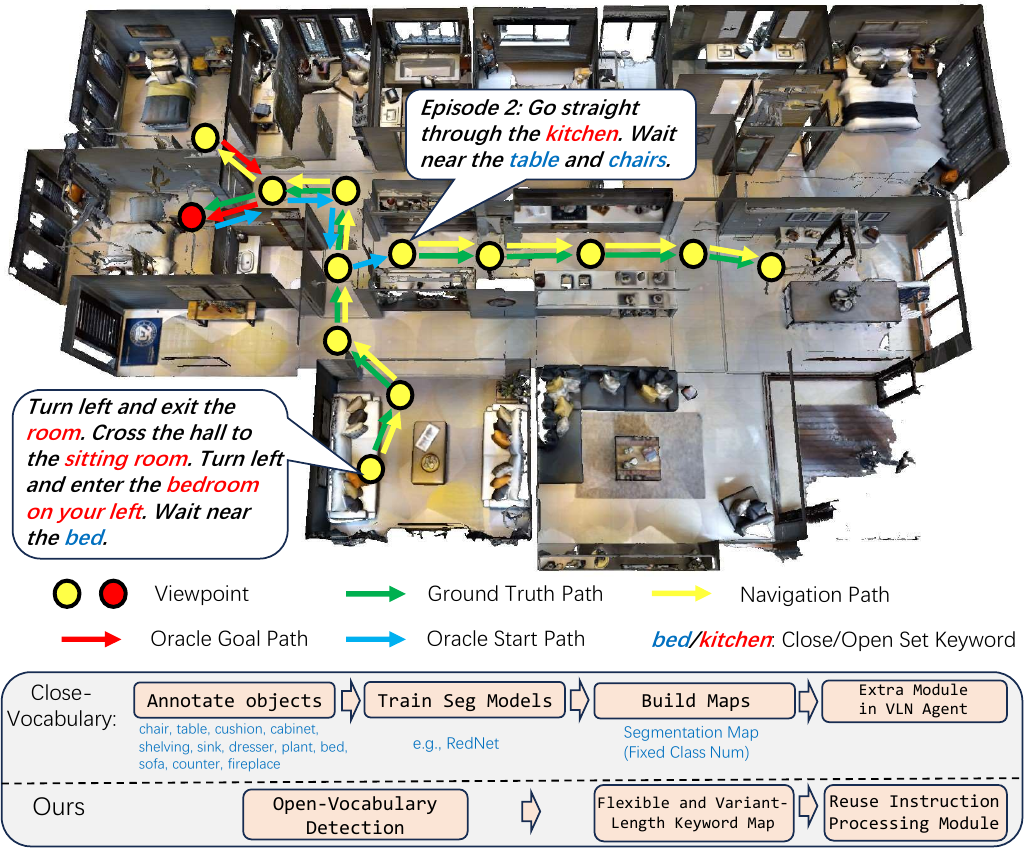}

   \caption{Top: Example of a two-episode tour. The agent first navigates the environment following the instruction of episode 1 (\textcolor{yellow}{Yellow}). Then the agent is directed to the ground truth goal as Oracle Goal phase (\textcolor{red}{Red}). Later the agent travels to the start point of episode 2 in the Oracle Start phase (\textcolor{blue}{Blue}). Finally, the agent navigates the environment following the next instruction (\textcolor{yellow}{Yellow}).
   Bottom: Comparison between previous methods and ours. Close-vocabulary methods require extra annotation and training efforts to provide segmentation results in navigation, and the agent is limited to a close set of categories building segmentation maps. Our method proposed an open-vocabulary-based omnigraph which is more flexible for various keywords and circumstances.}
   \label{fig:teaser}
\end{figure}


We demonstrate a typical procedure of an IVLN task in Figure~\ref{fig:teaser}. 
An IVLN agent receives a ordered sequence of instructions that fulfill a tour of the target scene.
Each tour consists of individual episodes, each guided by a language instruction.
When the agent completes the navigation of an instruction, it is teleoperated by an oracle to the correct goal location, and then translated to the start of the next episode. Hence, each episode is composed of three parts: navigation, oracle goal, and oracle start. 
In the navigation phase, the agent receives the environment observations along the path $P_I$ following the given instruction $I$. In the oracle goal/start phase, no instruction is present except the observations along the oracle start path $P_{OS}$ and the oracle goal path $P_{OG}$.
Thus the tour history of each episode is represented as $\{P_{OS}, (I, P_I), P_{OG}\}$.



Although the memory-retaining strategy aligns well with the persistent nature of the VLN tasks, it poses additional challenges on how to fully exploit the unstructured navigation history of previous episodes. 
First, it remains formidable to interpret the multi-modal information that spans a number of domains including language (instruction), vision (visual measurements), time, and location (physical movement), without any explicit supervision. 
The only weak supervision existing in the tour history is the correspondence between the instruction $I$ and the navigation path $P_I$. However, such correspondences are coarse and ambiguous as the visual observations along the path may not correspond to the order of the objects and actions appeared in the instruction.
Moreover, $P_I$ could deviate from the correct path due to the erroneous decisions made along the navigation.
Therefore, obtaining reliable supervision over the navigation history and establishing faithful correspondence between the multi-modal data are the keys to the success of IVLN tasks.

In the ideal scenarios where all the multi-modal data can be thoroughly interpreted, the second challenge arises in structurizing the extensive memory so that it can be effectively utilized under the IVLN framework. Prior works~\cite{krantz2023iterative} show that a naive stacking of the history data for feature representation would lead to inferior performance. Additionally, a well structured memory should be general enough to accommodate varying settings in IVLN tasks, e.g. the discrete~\cite{anderson2018vision,krantz2023iterative} and continuous environments~\cite{krantz2020beyond, krantz2023iterative} which are conventionally tackled with distinct strategies.  

To address the above challenges, we present \methodnamelong{}, a novel framework that strives to go over and beyond the current arts of IVLN solutions.
To combat with the lack of explicit supervisions, we propose to incorporate Large Language Models~(LLMs) and Open-Vocabulary Detection (OVD) to extract key information from the unattended data flow.
Specifically, while LLMs are leveraged to identify keywords from the language instructions, the OVD detectors are employed to build the correspondence between the keywords and the visual observations.
Such a strategy is capable of providing distilled supervision which is critical to understanding the unordered navigation data.
As the OVD detector can scale up to novel categories, \methodnamelong is able to generalize to unseen scenes on the fly without the need of extra annotation and re-training, offering great flexibility over the methods using closed-set detectors~\cite{krantz2023iterative,cartillier2021semantic}.  

To better harness the extracted supervisory signals, we further introduce a structured representation, coded \textit{Omnigraph}, to integrate multi-modal information along the tour.
By leveraging a novel fusion mechanism, omnigraph can efficiently collect the pertinent knowledge for a more accurate navigation action.
Moreover, the omnigraph representation can seamlessly support both discrete and continuous environments under a unified framework.
We extensively evaluate \methodnamelong on a number of challenging benchmarks.
The experimental results demonstrate the superiority of our method over the state-of-the-art approaches.
In summary, our contributions are:
\begin{itemize}
    \item A novel framework dubbed \methodnamelong, that, for the first time, incorporates LLMs and OVD into the IVLN paradigms to distill reliable and generalizable supervision signals from the unordered navigation data. 

    \item A structured and general representation called Omnigraph that facilitates the utilization of multi-modal knowledge and can be generalized to different VLN settings. 

    \item Superior performance on the IVLN tasks in both discrete and continuous environments. 

\end{itemize}

%% file: sec/2_related.tex
\section{Related Works}
\label{sec:related_works}

\noindent \textbf{Vision-and-Language Navigation}
Vision-and-Language Navigation~(VLN)~\cite{anderson2018vision,ku2020room,qi2020reverie,wang2019reinforced,fried2018speaker,kamath2023new,gao2023adaptive,huo2023geovln,li2023improving,li2023kerm,hwang2023meta} requires an agent with the ability to navigate a never-before-seen environment following a natural language instruction that describes the ground truth navigation path. There are two major settings in VLN benchmarks, discrete~\cite{anderson2018vision,ku2020room,qi2020reverie} and continuous environments~\cite{krantz2020beyond,savva2019habitat}. In the discrete setting, the VLN agent is limited to changing position and orientation by discrete amounts or predefined options, while the continuous setting provides a continuous range for the agent's action. Iterative VLN~\cite{krantz2023iterative} evaluates the agent in persistent environments, and the agent needs to utilize prior experience in the environment for better performance. 

\vspace{1mm}
\noindent\textbf{Persistent Environment and Iterative VLN}
The increasing amount and improved quality of 3D scene datasets~\cite{chang2017matterport3d,ramakrishnan2021habitat}, and the high-performance navigation environment simulation platform~\cite{kolve2017ai2,savva2019habitat,xia2018gibson,makoviychuk2021isaac} significantly promote the development of navigation tasks, and make it possible to study the long-horizon tasks in persistent environments such as visual navigation~\cite{wani2020multion,weihs2021visual}, multi-object navigation~\cite{wani2020multion}, visual room rearrangement~\cite{weihs2021visual}, courier task~\cite{mirowski2018learning}, multi-target embodied QA~\cite{yu2019multi}, scene exploration and object search~\cite{fang2019scene}.
Iterative VLN~\cite{krantz2023iterative} further studies VLN in persistent environments and enriches the long-horizon visual navigation problem with natural language and linguistic information.
Previous IVLN studies demonstrate that structured memory~\cite{cartillier2021semantic,kwon2023renderable,planche2019incremental,henriques2018mapnet,chaplot2020learning,georgakis2022learning,chen2019learning,ramakrishnan2020occupancy,wani2020multion} is essential for IVLN agents. TourHAMT~\cite{krantz2023iterative} tries to mitigate the problem in discrete environments by adding the history embedding from previous episodes to the current episode but fails to improve the performance. 
MAP-CMA~\cite{krantz2023iterative} constructs semantic and occupancy maps~\cite{cartillier2021semantic} from the point cloud built by fine-tuned RedNet~\cite{jiang2018rednet} and depth images for action prediction.  
However, point cloud construction requires depth sensors, and the RedNet is close-vocabulary and only limited to thirteen kinds of objects. This requires extra data collection labor for fine-tuning and prevents the agent from scaling to unseen environments and complicated concepts. 
TourHAMT/MAP-CMA can only be applied to discrete/continuous environments. Despite MAP-CMA's success, it is hard to transfer its solution to discrete environments as semantic maps require continuous observations.


\vspace{1mm}
\noindent\textbf{Open-Vocabulary Detection}
Open-vocabulary detection (OVD) aims to train object detectors beyond recognizing only base categories present in training labels and expand the vocabulary to detect novel categories. 
With the development of powerful language encoders~\cite{kenton2019bert} and contrastive image-text training~\cite{radford2021clip,zhai2022lit,jia2021scaling}, recent works transfer the language capabilities of these models to OVD~\cite{gu2022open,kamath2021mdetr,li2022grounded,zareian2021open,zhong2022regionclip,zhou2022detecting}. ViLD~\cite{gu2022open} distills the knowledge from a pretrained open-vocabulary image classification model into a two-stage detector following a teacher-student training paradigm. 
MDETR~\cite{kamath2021mdetr} and GLIP~\cite{li2022grounded} take a single text query for the image then formulate detection as the phrase grounding problem. Owl-ViT~\cite{minderer2022simple} combines Vision Transformer~\cite{dosovitskiy2021image}, contrastive image-text pre-training~\cite{radford2021clip} and end-to-end detection fine-tuning. We study OVD in persistent environments thus promoting the development of VLN.

%% file: sec/3_method.tex
\section{OVER-NAV}
\label{sec:method}

\begin{figure*}[t]
  \centering
   \includegraphics[width=0.95\linewidth]{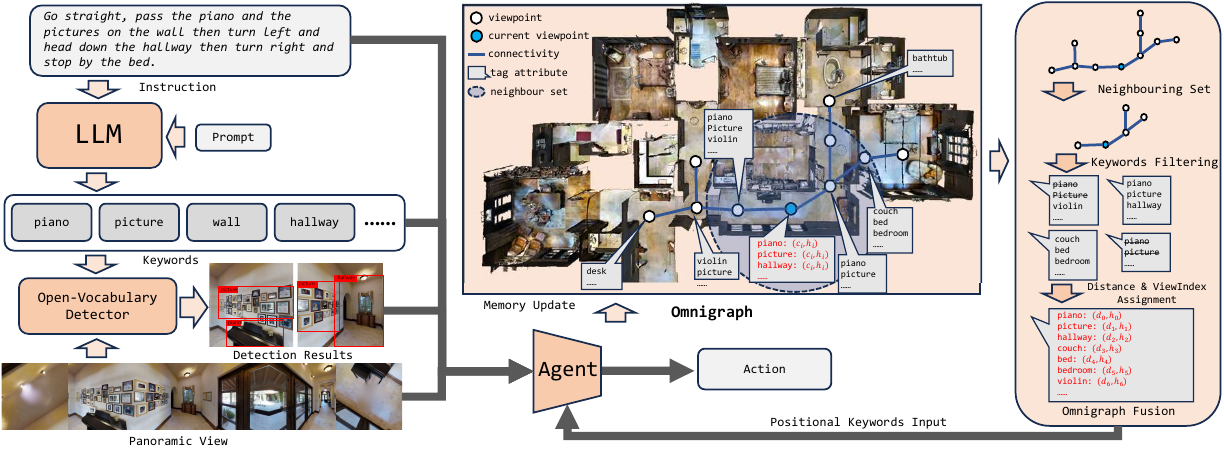}

   \caption{The overview of our proposed method. The instruction is sent to LLMs with the prompt to obtain keywords. The open-vocabulary detector receives the keywords and the panoramic view at the current position, and sends the detection results to the agent. With the detection results containing the distribution of detected objects, \eg, heading and confidence, the agent maintains the omnigraph that stores the information of visited viewpoints in previous episodes. Each viewpoint is tagged with keywords and distribution information. For inference, the omnigraph first collects the neighboring viewpoints and filters their keywords, then fuses the keywords with corresponding positional information, \eg, heading and confidence. Finally, the resulting positional keyword inputs are sent to the agent for prediction.}
   \label{fig:overview}
\end{figure*}

The framework of our proposed method is depicted in Fig~\ref{fig:overview}. 
Large Language Models~(LLMs) extract keywords from the instruction of each episode.
As the agent traverses the environment, it sends the keywords and observations along its path to the OVD detector for detection. The detection results are subsequently stored in the \keywordmap, which is established and maintained to memorize the observed portion of the current environment. The \keywordmap generates the keyword input for the agent's action prediction. This input encapsulates the information of an ego-centric map, thereby aiding the prediction process.


\subsection{OVD-based Omnigraph Construction}

The \methodnamelong aims to construct an organized \keywordmap that allows the IVLN agent to memorize and utilize the history information from previous episodes in the same tour. It incorporates three distinct processes: keyword extraction, keyword-panoramic detection, and \keywordmap construction.

\vspace{1mm}


\noindent\textbf{Keyword Extraction} The IVLN agent receives an instruction each episode in the tour.
A typical navigation instruction can be decomposed into two parts: milestones and actions. The milestones~(hereinafter referred to as keywords) are related to the scene where the navigation task is issued. Conversely, actions are likely episodic and related to the agent's current orientation. 
For instance, consider the instruction \textit{``Head past the dining table and turn left towards the kitchen''}, the terms \textit{dining table} and \textit{kitchen} serve as milestones, enabling the agent to discern different directions and movements (\eg, \textit{head past}, \textit{turn left}) and verify the accuracy of previous actions.
Thus the keywords provide the agent with a condensed understanding of the scene.

We propose employing Large Language Models~(LLMs)~\cite{openai2023gpt4, brown2020language} for keyword extraction. The system prompt of GPT is provided in supplementary material. The GPT is fed instructions following the format defined in the prompt. Subsequently, the LLM responds to the query with the appropriately formatted keywords. One of the key benefits of using LLMs for keyword extraction lies in their flexibility because the keywords in instructions can vary greatly in length and category. 
For instance, keywords like ``\textit{counter with the blue top}'' carry significant attributes for the agent's reference, and LLMs can effectively identify such keywords with their strong in-context learning ability.

\vspace{1mm}

\noindent \textbf{Keyword-Panoramic Detection} After keyword extraction, the agent extracts a keyword set $K_I = \{k_i\}_{i=1}^N$ with $N$ keywords from each instruction $I$. 
Throughout the navigation process of instruction $I$, the agent constantly observes the environment and captures images from its current position and orientation.
Following the common practice of some VLN models, \eg, HAMT~\cite{chen2021history}, 
we assume that the agent can acquire panoramic photos $P_{pano}$ of the surrounding environment. This can be achieved by equipping the agent with a panoramic camera or rotating the agent 360$^{\circ}$.

Upon reaching a position and capturing panoramic images, the agent can perform open-vocabulary detection using the keyword set $K_I$ and the image $P_{pano}$. This position is subsequently recorded as a \textbf{viewpoint} in the agent's memory. The OVD detector $D$ generates a set of detection boxes $\{B_i, l_i, c_i\}_{i=1}^{M}$ as detection results, where each box $B_i$ is with a label $l_i$ and a confidence score $c_i$. Each box $B_i$ contains four values $(x_{min}, y_{min}, x_{max}, y_{max})$ to locate the object in the panoramic image. These values can be used to calculate the relative heading of the detected object concerning the agent's orientation. This can then be transformed into the absolute heading $h_i$ in the coordinate system of the current environment by deducting the agent's heading. Thus, the detection results stored in the agent's memory for the current viewpoint are $\{B_i, l_i, c_i, h_i\}_{i=1}^{M}$.

In each episode, the agent performs the detection with $K_I$ and $P_{pano}$ at every viewpoint along the navigation path. 
Each viewpoint along the path is tagged with detection boxes, labels, confidence, and absolute heading, which summarize the properties of viewpoints in the scene. We will further elaborate the details, \eg, the location of viewpoints, in Sec.~\ref{sec:hamt_ours} and Sec.~\ref{sec:mapcma_ours}.


\vspace{1mm}

\noindent\textbf{\KeywordMap Construction} After keyword extraction and keyword-panoramic detection, all visited viewpoints are tagged with keywords. We then construct the \textbf{\keywordmap} as a graph whose nodes are the viewpoints with keywords and edges are the connectivity between viewpoints. The \keywordmap organizes the viewpoint, delineates the structure of the current scene, and outlines the distribution of various objects identified through panoramic keyword detection. It can be readily applied to both discrete and continuous environments. Furthermore, the open-vocabulary \keywordmap can handle a wide array of keywords, and also empowers the model to discern the intrinsic relationship between keywords. For example, if the agent determines the relationship between the \textit{living room} and the \textit{television}, it can make an informed decision when instructed to go to the \textit{living room}, even if only the \textit{television} is present in the \keywordmap.


Specifically, when the agent arrives at a new viewpoint $a$ from a previous viewpoint $b$, and obtains the detection results, it incorporates the viewpoint $a$ into the graph as a new node, and adds the undirected edge $\langle a, b\rangle$ to the \keywordmap. If $a$ is the starting point, only $a$ is added to the graph. During the navigation, the \keywordmap is incrementally refined in two ways: i) the discovery of viewpoints updates the nodes and edges, ii) when the agent arrives at a previously visited viewpoint discovered in an earlier episode $E_b$ in a new episode $E_a$, the keyword-panoramic detection will generate fresh detection results with the new instruction $I_{E_a}$, which might have different keywords from the previous instruction $I_{E_b}$. The new detection results are then employed to update the keywords associated with this viewpoint.

\vspace{1mm}




\subsection{Omnigraph Fusion}

After the \keywordmap construction, the agent needs to exploit the information in the \keywordmap at each step during its navigation. 
However, directly feeding the entire \keywordmap to the agent is computationally intensive and inefficient as the agent only navigates a small area of the entire scene and the number of detection boxes might be overwhelmingly large. Hence, we fuse the information of the omnigraph within a local and ego-centric subgraph and send it to the agent. We reuse the agent's instruction encoder to extract the embeddings of keywords in \keywordmap subgraph, which not only eliminates the cost of the extra module but also preserves the semantic consistency between instruction embeddings and keyword embeddings.
Then the extra information for each keyword is fused to enrich the keyword embeddings.
Here we discuss the omnigraph fusion in discrete and continuous environments respectively. The overviews of the two agents are provided in supplementary materials.

\subsubsection{\methodnameshort for Discrete Environments}
\label{sec:hamt_ours}

In discrete VLN, the environment is represented as a set of pre-defined viewpoints, and the agent can navigate through the connections between these viewpoints. Following previous studies, we integrate \methodnameshort into the History Aware Multimodal Transformer~(HAMT)~\cite{chen2021history} framework to work with discrete environments.

HAMT~\cite{chen2021history} trains the agent with both text-modal and vision-modal inputs. HAMT agent encodes the input instruction $\mathcal{W} = (w_1, w_2, ... , w_L)$ with $L$ words as instruction embeddings $X = (x_{CLS}, x_1, x_2, ... , x_L)$ using BERT, and encodes the panoramic observation $\mathcal{O}_t = (v_1^o, v_2^o, ... , v_K^o)$ at step $t$ with $K$ different views as observation embeddings $O_t = (o_1, o_2, ..., o_K, o_{stop})$. HAMT proposes hierarchical history encoding to encode the navigation history $\mathcal{H}_t$ at step $t$, which consists of all the past panoramic observations $\mathcal{O}_{1,...,t-1}$ and performed actions $a_{1,...,t-1}$ before step $t$. The hierarchical history encoding produces history embedding $H_t = (h_{CLS}, h_1, ... , h_{t-1})$. Then the history embedding and observation embedding are concatenated as vision modality, while instruction embedding serves as text modality. The cross-modal transformer fuses them and outputs the probability distribution of selecting different views of observation $O_t$ as the predicted action.

Since the discrete environment is already discretized into viewpoints, the \methodnameshort agent performs object detection at each new viewpoint encountered, \ie, at every step. For an agent positioned at viewpoint $p$, we prepare the omnigraph input using the following steps:

\begin{enumerate}
    \item Neighbours Identification. Given distance $d_n$, we collect the neighboring viewpoints that are reachable from the current position within $d_n$ steps as the neighbor set. 

    \item Inner-viewpoint Detection Box Filtering. For each neighbor, we filter out most of the detection boxes so that each keyword detected in this neighbor is associated with only one detection box with the highest detection confidence $c_i$.

    \item Distance \& View Index Assignment. For keyword $k$ of neighbouring viewpoint $v$, we assign the distance $d_v^k$ between $v$ and the current position to $k$, and the view index $h_v^k$ which is the direction of the next move the agent should take to go to $v$ from the current position to $k$. Please note that both $d_v^k$ and $h_v^k$ are discrete rather than continuous, which makes the next step possible.
    
    \item Cross-viewpoint Keyword Filtering. For those keywords that appear in more than one neighbour and thus have multiple $d_v^k$ and $h_v^k$, we select the most frequent $d_v^k$ and $h_v^k$ as final attributes for $k$.
\end{enumerate}

After these procedures, the agent obtains a set of distinct keywords and each keyword has a distance $d_v^k$ and a view index $h_v^k$. Following HAMT, we use BERT to extract the embeddings of keywords, then use the [CLS] token embeddings $E_{CLS}$ as the representation of each keyword. The view index $h_v^k$ is converted to 4-dimension embeddings $E_{h_v^k} = (\sin\theta, \cos\theta, \sin\phi, \cos\phi)$ where $\theta$ and $\phi$ are the heading and elevation of view index $h_v$. Therefore, the fused keyword embedding is computed as:
\begin{equation}
    E_{k} = LN(W_{CLS} E_{CLS}) + LN(W_{h_v^k}E_{h_v^k}) + E_{d_v^k},
\end{equation}
where $E_{d_v^k}$ is the distance embedding, $W_{CLS}$ and $W_{h_v^k}$ are learnable weights, and $LN$ is layer normalization. Then all embeddings of keywords are concatenated:
\begin{equation}
    E_{map} = \mathrm{Concat}(E_{k_1}, E_{k_2}, ... , E_{k_N}),
\end{equation}
where $N$ is the number of keywords and $E_{k_i}$ is the fused keyword embedding of the $i$-th closest keyword $k_i$ to the agent's current position.

The embedding 
$E_{map}$ can be viewed as the text-modal context that indicates the position of every detected object in the memory. Therefore, we concatenate $E_{map}$ to instruction embeddings $X$ and send them to the agent as text modality.

\setlength{\tabcolsep}{.3em}
\begin{table*}[t]
    \renewcommand{\arraystretch}{1.15}
    \setlength{\aboverulesep}{0pt}
    \setlength{\belowrulesep}{0pt}
    \centering
	\resizebox{0.9\textwidth}{!}{
		\begin{tabular}{clcccc c ccccccs c ccccccs}
			\toprule
            & & & & & &
			& \multicolumn{7}{c}{\scriptsize\textbf{Val-Seen}}
		   && \multicolumn{7}{c}{\scriptsize\textbf{Val-Unseen}}
            \\
			\cmidrule{8-14}
			\cmidrule{16-22}
			\scriptsize \shortstack{\#} &
			{\scriptsize Model}
			& \footnotesize \textsc{ph}
			& \footnotesize \textsc{th}
			& \footnotesize \textsc{phi}
			& \footnotesize \textsc{iw}
			&
			& \scriptsize\textbf{\texttt{TL}}
			& \scriptsize\textbf{\texttt{NE}}~$\downarrow$
			& \scriptsize\textbf{\texttt{OS}}~$\uparrow$
			& \scriptsize\textbf{\texttt{nDTW}}~$\uparrow$
			& \scriptsize\textbf{\texttt{SR}}~$\uparrow$
			& \scriptsize\textbf{\texttt{SPL}}~$\uparrow$
			& \scriptsize\textbf{\texttt{t-nDTW}}~$\uparrow$
			&
			& \scriptsize\textbf{\texttt{TL}}
			& \scriptsize\textbf{\texttt{NE}}~$\downarrow$
			& \scriptsize\textbf{\texttt{OS}}~$\uparrow$
			& \scriptsize\textbf{\texttt{nDTW}}~$\uparrow$
			& \scriptsize\textbf{\texttt{SR}}~$\uparrow$
			& \scriptsize\textbf{\texttt{SPL}}~$\uparrow$
			& \scriptsize\textbf{\texttt{t-nDTW}}~$\uparrow$
			\\
			\midrule
			\scriptsize \texttt{1}
                & \texttt{HAMT}
                & & & & &
                & 10.1 \scriptsize{$\pm$0.1}
                & 4.2 \scriptsize{$\pm$0.1}
                & \bf 70 \scriptsize{$\pm$1}
                & 71 \scriptsize{$\pm$1}
                & 63 \scriptsize{$\pm$1}
                & 61 \scriptsize{$\pm$1}
                & 58 \scriptsize{$\pm$1}
                &
                & \hphantom{0}9.4 \scriptsize{$\pm$0.1}
                & 4.7 \scriptsize{$\pm$0.0}
                & 64 \scriptsize{$\pm$1}
                & 66 \scriptsize{$\pm$0}
                & 56 \scriptsize{$\pm$0}
                & 54 \scriptsize{$\pm$0}
                & 50 \scriptsize{$\pm$0}
			\\
			\scriptsize \texttt{2}
			    & \texttt{TourHAMT}
			    & \checkmark & \checkmark & \checkmark & \checkmark &
			    & \hphantom{0}9.4 \scriptsize{$\pm$0.4}
                & 5.8 \scriptsize{$\pm$0.1}
                & 56 \scriptsize{$\pm$1}
                & 59 \scriptsize{$\pm$0}
                & 45 \scriptsize{$\pm$1}
                & 43 \scriptsize{$\pm$1}
                & 45 \scriptsize{$\pm$0}
                &
                & 10.0 \scriptsize{$\pm$0.2}
                & 6.2 \scriptsize{$\pm$0.1}
                & 52 \scriptsize{$\pm$2}
                & 52 \scriptsize{$\pm$0}
                & 39 \scriptsize{$\pm$1}
                & 36 \scriptsize{$\pm$0}
                & 32 \scriptsize{$\pm$1}
			\\
			\scriptsize \texttt{3}
			    & 
			    & \checkmark & \checkmark & \checkmark &  &
                & 10.5 \scriptsize{$\pm$0.3}
                & 6.0 \scriptsize{$\pm$0.2}
                & 60 \scriptsize{$\pm$1}
                & 58 \scriptsize{$\pm$1}
                & 45 \scriptsize{$\pm$2}
                & 43 \scriptsize{$\pm$2}
                & 42 \scriptsize{$\pm$1}
                &
                & 10.9 \scriptsize{$\pm$0.2}
                & 6.8 \scriptsize{$\pm$0.2}
                & 54 \scriptsize{$\pm$1}
                & 51 \scriptsize{$\pm$1}
                & 38 \scriptsize{$\pm$1}
                & 34 \scriptsize{$\pm$1}
                & 31 \scriptsize{$\pm$1}
			\\
			\scriptsize \texttt{4}
			    & 
			    & \checkmark & \checkmark &  &  &
                & 10.6 \scriptsize{$\pm$0.3}
                & 6.0 \scriptsize{$\pm$0.1}
                & 61 \scriptsize{$\pm$1}
                & 58 \scriptsize{$\pm$1}
                & 45 \scriptsize{$\pm$1}
                & 42 \scriptsize{$\pm$1}
                & 42 \scriptsize{$\pm$1}
                &
                & 10.3 \scriptsize{$\pm$0.3}
                & 6.7 \scriptsize{$\pm$0.2}
                & 52 \scriptsize{$\pm$1}
                & 50 \scriptsize{$\pm$1}
                & 38 \scriptsize{$\pm$1}
                & 34 \scriptsize{$\pm$1}
                & 29 \scriptsize{$\pm$1}
			\\
			\scriptsize \texttt{5}
			    & 
			    & \checkmark &  &  &  &
                & 10.9 \scriptsize{$\pm$0.3}
                & 6.1 \scriptsize{$\pm$0.1}
                & 60 \scriptsize{$\pm$2}
                & 58 \scriptsize{$\pm$1}
                & 45 \scriptsize{$\pm$1}
                & 42 \scriptsize{$\pm$1}
                & 41 \scriptsize{$\pm$0}
                &
                & 11.0 \scriptsize{$\pm$0.6}
                & 6.7 \scriptsize{$\pm$0.1}
                & 52 \scriptsize{$\pm$2}
                & 51 \scriptsize{$\pm$0}
                & 38 \scriptsize{$\pm$0}
                & 34 \scriptsize{$\pm$0}
                & 28 \scriptsize{$\pm$1}
			\\
                \midrule
                \scriptsize \texttt{6}
                & \texttt{Ours}
                & & & & &
                & 9.9 \scriptsize{$\pm$0.1}
                & \bf 3.7 \scriptsize{$\pm$0.1}
                & \bf 70 \scriptsize{$\pm$0}
                & \bf 73 \scriptsize{$\pm$1}
                & \bf 65 \scriptsize{$\pm$1}
                & \bf 63 \scriptsize{$\pm$1}
                & \bf 62 \scriptsize{$\pm$0}
                &
                & \hphantom{0}9.4 \scriptsize{$\pm$0.1}
                & \bf 4.1 \scriptsize{$\pm$0.1}
                & \bf 66 \scriptsize{$\pm$1}
                & \bf 69 \scriptsize{$\pm$0}
                & \bf 60 \scriptsize{$\pm$1}
                & \bf 57 \scriptsize{$\pm$0}
                & \bf 55 \scriptsize{$\pm$1}
			\\
			\bottomrule
		\end{tabular}}
	\caption{
	    The comparison between our method, HAMT and \thamt on \benchmark. 
	    \textsc{ph}: previous episodes' history; \textsc{th}: trainable history encoder; \textsc{phi}: previous history identifier; \textsc{iw}: inflection weighting. \thamt fails to outperform HAMT, while our method achieves significant improvements in both val-seen and val-unseen datasets. We run each experiment 3 times and report metrics as $\bar{x} \pm \sigma_{\bar{x}}$.
	}
	\label{tab:ivln_results}
\end{table*}

\setlength{\tabcolsep}{.3em}
\begin{table*}[t]
    \setlength{\aboverulesep}{0pt}
    \setlength{\belowrulesep}{0pt}
    \renewcommand{\arraystretch}{1.15}
    \centering
	\resizebox{0.8\textwidth}{!}{
		\begin{tabular}{cl c cccccs c ccccccs}
			\toprule
            & 
			& \multicolumn{7}{c}{\scriptsize\textbf{Val-Seen}}
		   && \multicolumn{7}{c}{\scriptsize\textbf{Val-Unseen}}
            \\
			\cmidrule{3-9}
			\cmidrule{11-17}
			\scriptsize \shortstack{\#} &
			{\scriptsize Model}
			& \scriptsize\textbf{\texttt{TL}}
			& \scriptsize\textbf{\texttt{NE}}~$\downarrow$
			& \scriptsize\textbf{\texttt{OS}}~$\uparrow$
			& \scriptsize\textbf{\texttt{nDTW}}~$\uparrow$
			& \scriptsize\textbf{\texttt{SR}}~$\uparrow$
			& \scriptsize\textbf{\texttt{SPL}}~$\uparrow$
			& \scriptsize\textbf{\texttt{t-nDTW}}~$\uparrow$
			&
			& \scriptsize\textbf{\texttt{TL}}
			& \scriptsize\textbf{\texttt{NE}}~$\downarrow$
			& \scriptsize\textbf{\texttt{OS}}~$\uparrow$
			& \scriptsize\textbf{\texttt{nDTW}}~$\uparrow$
			& \scriptsize\textbf{\texttt{SR}}~$\uparrow$
			& \scriptsize\textbf{\texttt{SPL}}~$\uparrow$
			& \scriptsize\textbf{\texttt{t-nDTW}}~$\uparrow$
			\\
			\midrule
			\scriptsize \texttt{1}
                & \texttt{CMA}
                &  7.8 \scriptsize{$\pm$0.4}
                &  8.8 \scriptsize{$\pm$0.6}
                & 27   \scriptsize{$\pm$3 }
                & 42   \scriptsize{$\pm$3 }
                & 18   \scriptsize{$\pm$3 }
                & 17   \scriptsize{$\pm$3 }
                & 39   \scriptsize{$\pm$1 }
                &&  7.5 \scriptsize{$\pm$0.3}
                 &  8.8 \scriptsize{$\pm$0.2}
                 & 26   \scriptsize{$\pm$1 }
                 & 44   \scriptsize{$\pm$1 }
                 & 19   \scriptsize{$\pm$1 }
                 & 18   \scriptsize{$\pm$1 }
                 & 38   \scriptsize{$\pm$2 }
			\\
			\scriptsize \texttt{2}
			    & \texttt{TourCMA}
                &  8.0 \scriptsize{$\pm$0.4}
                &  8.2 \scriptsize{$\pm$0.9}
                & 30   \scriptsize{$\pm$2 }
                & 44   \scriptsize{$\pm$2 }
                & 20   \scriptsize{$\pm$3 }
                & 19   \scriptsize{$\pm$2 }
                & 40   \scriptsize{$\pm$1 }
                &&  7.8 \scriptsize{$\pm$0.1}
                 &  9.0 \scriptsize{$\pm$0.2}
                 & 26   \scriptsize{$\pm$1 }
                 & 42   \scriptsize{$\pm$1 }
                 & 18   \scriptsize{$\pm$0 }
                 & 17   \scriptsize{$\pm$1 }
                 & 36   \scriptsize{$\pm$1 }
			\\
			\scriptsize \texttt{3}
			    & \texttt{PoolCMA}
                &  7.2 \scriptsize{$\pm$0.5}
                &  9.1 \scriptsize{$\pm$0.4}
                & 24   \scriptsize{$\pm$4 }
                & 41   \scriptsize{$\pm$2 }
                & 17   \scriptsize{$\pm$4 }
                & 16   \scriptsize{$\pm$2 }
                & 37   \scriptsize{$\pm$2 }
                &&  7.3 \scriptsize{$\pm$0.2}
                 &  9.0 \scriptsize{$\pm$0.3}
                 & 23   \scriptsize{$\pm$1 }
                 & 42   \scriptsize{$\pm$1 }
                 & 16   \scriptsize{$\pm$1 }
                 & 15   \scriptsize{$\pm$0 }
                 & 36   \scriptsize{$\pm$2 }
			\\
			\scriptsize \texttt{4}
			    & \texttt{PoolEndCMA}
                &  7.6 \scriptsize{$\pm$0.8}
                &  8.9 \scriptsize{$\pm$0.9}
                & 27   \scriptsize{$\pm$3 }
                & 42   \scriptsize{$\pm$3 }
                & 18   \scriptsize{$\pm$4 }
                & 17   \scriptsize{$\pm$2 }
                & 38   \scriptsize{$\pm$2 }
                &&  6.9 \scriptsize{$\pm$0.2}
                 &  8.7 \scriptsize{$\pm$0.2}
                 & 25   \scriptsize{$\pm$2 }
                 & 44   \scriptsize{$\pm$1 }
                 & 18   \scriptsize{$\pm$1 }
                 & 16   \scriptsize{$\pm$1 }
                 & 38   \scriptsize{$\pm$2 }
			\\
                \scriptsize \texttt{5}
			    & \texttt{MAP-CMA}
                &  9.4
                &  6.4
                & 48
                & 56
                & \textbf{39}
                & \textbf{36}
                & 52
                &&  8.5
                 &  6.8
                 & 44
                 & 54
                 & \bf 35
                 & 32
                 & 47
			\\
                \midrule
                \scriptsize \texttt{6}
			    & \texttt{Ours}
                &  9.5 \scriptsize{$\pm$0.9}
                & \bf 5.8 \scriptsize{$\pm$0.9}
                & \bf 49   \scriptsize{$\pm$4 }
                & \bf 59   \scriptsize{$\pm$2 }
                & \bf 39   \scriptsize{$\pm$2 }
                & \bf 36   \scriptsize{$\pm$2 }
                & \bf 56   \scriptsize{$\pm$2 }
                &&  8.8 \scriptsize{$\pm$0.6}
                 &  \bf 6.5 \scriptsize{$\pm$0.2}
                 & \bf 45   \scriptsize{$\pm$2 }
                 & \bf 56   \scriptsize{$\pm$1 }
                 & \bf 35   \scriptsize{$\pm$1 }
                 & \bf 33   \scriptsize{$\pm$1 }
                 & \bf 50   \scriptsize{$\pm$2 }
			\\
			\bottomrule
		\end{tabular}
        }
	\caption{The performance of our method on IR2R-CE. The experiment results of all previous methods are copied from ~\cite{krantz2023iterative}. Our method and MAP-CMA both use inferred semantics and iterative map construction. Our method gains 4\% and 3\% t-nDTW improvement on val-seen and val-unseen datasets respectively. We run each experiment 3 times and report metrics as $\bar{x} \pm \sigma_{\bar{x}}$.
	}
	\label{tab:ivlnce_results}
\end{table*}

\subsubsection{\methodnameshort for Continuous Environments}
\label{sec:mapcma_ours}
VLN in continuous environments discards the pre-defined viewpoints for navigation and situates the agent in continuous environments with low-level actions. Following previous studies, we apply our \methodnameshort to continuous environments by incorporating \methodnameshort into a variant of Cross-Modal Attention~\cite{krantz2020beyond}, MAP-CMA~\cite{krantz2023iterative}.

MAP-CMA~\cite{krantz2023iterative} is based on CMA~\cite{krantz2020beyond}, a common baseline in recent works. CMA is an end-to-end recurrent model that predicts actions from RGBD observations, the instruction and previous actions. CMA uses two recurrent networks, one for visual history tracking, and the other for instruction and visual features. MAP-CMA replace the RGBD observations with occupancy maps and semantic maps, which are built by using inverse pinhole camera projection model to 3D pointclouds. MAP-CMA use a RedNet~\cite{jiang2018rednet} feature encoder to form a semantic pointcloud, which is fine-tuned on thirteen common labels.

The application of \methodnameshort to MAP-CMA is similar to HAMT, except that \methodnameshort needs to decide the position of viewpoints, \ie, where the keyword-panoramic detection should be performed. For an agent at position $p$, we prepare the omnigraph input with the following steps:
\begin{enumerate}
    \item Viewpoint Discovery. Given discovery threshold $d_{vp}$ as a hyper-parameter, when the agent arrives at a position whose distances to all recorded viewpoints in the memory are larger than $d_{vp}$, this position is registered as a new viewpoint $v$ and the agent stores the observations $P_{pano}^v$. 
    Different from Discrete VLN, we perform the keyword-panoramic detection with the stored observations $P_{pano}^v$ at viewpoint $v$ and keywords from current instruction when the agent arrives at a position whose distance to $v$ is less than the hyper-parameter viewpoint detection threshold $d_{det}$.
    $d_{det} < d_{vp}$. 

    \item Neighbours Identification. Given a hyper-parameter $d_n$, we collect the neighboring viewpoints whose distance to the current position is less than $d_n$.

    \item Inner-viewpoint Detection Box Filtering. This part is the same as Sec.~\ref{sec:hamt_ours}.

    \item Distance \& Heading Assignment. Similar to Sec.~\ref{sec:hamt_ours}, we assign heading $h_v^k$ and distance $d_v^k$ to each keyword.
    In this case, $d_v^k$ and $h_v^k$ are continuous.

    \item Cross-viewpoint Keyword Filtering. We select the detection box with the highest score for each keyword.
\end{enumerate}

Similar to Sec.~\ref{sec:hamt_ours}, the agent receives a set of distinct keywords and each keyword has a heading $h_v^k$ and distance $d_v^k$. We use the instruction encoder of MAP-CMA to extract the embedding of every keyword. Then we fuse the keyword with heading and distance embeddings using linear transformation. Finally, we perform the attention operation with the instruction embedding as query and the fused keyword embedding as key/value. The operation result of the attention is sent to the model for action prediction.

%% file: sec/4_experiment.tex
\section{Experiments}

\begin{table*}[t]
    \renewcommand{\arraystretch}{1.15}
    \setlength{\aboverulesep}{0pt}
    \setlength{\belowrulesep}{0pt}
    \centering
	\resizebox{0.8\textwidth}{!}{
		\begin{tabular}{cl c ccccccs c ccccccs}
			\toprule
            & &
			& \multicolumn{7}{c}{\scriptsize\textbf{Val-Seen}}
		   && \multicolumn{7}{c}{\scriptsize\textbf{Val-Unseen}}
            \\
			\cmidrule{4-10}
			\cmidrule{12-18}
			\scriptsize \shortstack{\#} &
			{\scriptsize Model}
			&
			& \scriptsize\textbf{\texttt{TL}}
			& \scriptsize\textbf{\texttt{NE}}~$\downarrow$
			& \scriptsize\textbf{\texttt{OS}}~$\uparrow$
			& \scriptsize\textbf{\texttt{nDTW}}~$\uparrow$
			& \scriptsize\textbf{\texttt{SR}}~$\uparrow$
			& \scriptsize\textbf{\texttt{SPL}}~$\uparrow$
			& \scriptsize\textbf{\texttt{t-nDTW}}~$\uparrow$
			&
			& \scriptsize\textbf{\texttt{TL}}
			& \scriptsize\textbf{\texttt{NE}}~$\downarrow$
			& \scriptsize\textbf{\texttt{OS}}~$\uparrow$
			& \scriptsize\textbf{\texttt{nDTW}}~$\uparrow$
			& \scriptsize\textbf{\texttt{SR}}~$\uparrow$
			& \scriptsize\textbf{\texttt{SPL}}~$\uparrow$
			& \scriptsize\textbf{\texttt{t-nDTW}}~$\uparrow$
			\\
			\midrule
			\scriptsize \texttt{1}
                & \texttt{Baseline}
                &
                & 10.1 \scriptsize{$\pm$0.1}
                & 4.2 \scriptsize{$\pm$0.1}
                & 70 \scriptsize{$\pm$1}
                & 71 \scriptsize{$\pm$1}
                & 63 \scriptsize{$\pm$1}
                & 61 \scriptsize{$\pm$1}
                & 58 \scriptsize{$\pm$1}
                &
                & \hphantom{0}9.4 \scriptsize{$\pm$0.1}
                & 4.7 \scriptsize{$\pm$0.0}
                & 64 \scriptsize{$\pm$1}
                & 66 \scriptsize{$\pm$0}
                & 56 \scriptsize{$\pm$0}
                & 54 \scriptsize{$\pm$0}
                & 50 \scriptsize{$\pm$0}
			\\
                \midrule
                \scriptsize \texttt{6}
                & \texttt{Type-I}
                &
                & 10.1 \scriptsize{$\pm$0.2}
                & 3.8 \scriptsize{$\pm$0.1}
                & 70 \scriptsize{$\pm$1}
                & 73 \scriptsize{$\pm$1}
                & 65 \scriptsize{$\pm$1}
                & 62 \scriptsize{$\pm$1}
                & 61 \scriptsize{$\pm$1}
                &
                & \hphantom{0}9.7 \scriptsize{$\pm$0.3}
                & 4.2 \scriptsize{$\pm$0.1}
                & 66 \scriptsize{$\pm$1}
                & 68 \scriptsize{$\pm$0}
                & 59 \scriptsize{$\pm$1}
                & 56 \scriptsize{$\pm$0}
                & 52 \scriptsize{$\pm$1}
			\\
                \scriptsize \texttt{6}
                & \texttt{Type-II}
                &
                & 10.2 \scriptsize{$\pm$0.3}
                & 3.7 \scriptsize{$\pm$0.1}
                & 71 \scriptsize{$\pm$1}
                & 73 \scriptsize{$\pm$1}
                & 65 \scriptsize{$\pm$1}
                & 62 \scriptsize{$\pm$1}
                & 62 \scriptsize{$\pm$0}
                &
                & \hphantom{0}9.8 \scriptsize{$\pm$0.2}
                & 4.2 \scriptsize{$\pm$0.2}
                & 66 \scriptsize{$\pm$1}
                & 68 \scriptsize{$\pm$1}
                & 59 \scriptsize{$\pm$1}
                & 56 \scriptsize{$\pm$0}
                & 53 \scriptsize{$\pm$1}
			\\
                \midrule
                \scriptsize \texttt{6}
                & \texttt{Ours}
                &
                & 9.9 \scriptsize{$\pm$0.1}
                & 3.7 \scriptsize{$\pm$0.1}
                & 70 \scriptsize{$\pm$0}
                & 73 \scriptsize{$\pm$1}
                & 65 \scriptsize{$\pm$1}
                & 63 \scriptsize{$\pm$1}
                & 62 \scriptsize{$\pm$0}
                &
                & \hphantom{0}9.4 \scriptsize{$\pm$0.1}
                & 4.1 \scriptsize{$\pm$0.1}
                & 66 \scriptsize{$\pm$1}
                & 69 \scriptsize{$\pm$0}
                & 60 \scriptsize{$\pm$1}
                & 57 \scriptsize{$\pm$0}
                & 55 \scriptsize{$\pm$1}
			\\
                \midrule
                \scriptsize \texttt{6}
                & \texttt{Type-III}
                &
                & 9.9 \scriptsize{$\pm$0.2}
                & 3.6 \scriptsize{$\pm$0.1}
                & 70 \scriptsize{$\pm$0}
                & 74 \scriptsize{$\pm$1}
                & 65 \scriptsize{$\pm$0}
                & 63 \scriptsize{$\pm$1}
                & 63 \scriptsize{$\pm$0}
                &
                & \hphantom{0}9.4 \scriptsize{$\pm$0.2}
                & 4.0 \scriptsize{$\pm$0.0}
                & 68 \scriptsize{$\pm$1}
                & 69 \scriptsize{$\pm$1}
                & 61 \scriptsize{$\pm$1}
                & 58 \scriptsize{$\pm$1}
                & 56 \scriptsize{$\pm$1}
			\\
			\bottomrule
		\end{tabular}}
	\caption{
        The ablation of open-vocabulary keywords. Type-I, Type-II and Type-III adopt different keyword strategies compared to our method. Type-I limits the keywords to 12 categories and obtains minimal improvement among these models. Type-II retains the attributes of the 12 keywords and slightly improves the performance of Type-I. Ours retains all open-vocabulary keywords and further gains 2\%. Type-III's memory contains the detection results in all viewpoints for the keywords of the current instruction in advance, thus achieving the best performance.
	}
	\label{tab:open_voca_ablation}
\end{table*}

\setlength{\tabcolsep}{.3em}
\begin{table}[t]
    \setlength{\aboverulesep}{0pt}
    \setlength{\belowrulesep}{0pt}
    \renewcommand{\arraystretch}{1.15}
    \centering
	\resizebox{0.6\linewidth}{!}{
		\begin{tabular}{cl c s c cs}
			\toprule
            & 
			& \multicolumn{2}{c}{\scriptsize\textbf{Val-Seen}}
		   && \multicolumn{2}{c}{\scriptsize\textbf{Val-Unseen}}
            \\
			\cmidrule{3-4}
			\cmidrule{6-7}
			\scriptsize \shortstack{\#} &
			{\scriptsize Model}
			& \scriptsize\textbf{\texttt{nDTW}}~$\uparrow$
			& \scriptsize\textbf{\texttt{t-nDTW}}~$\uparrow$
			&
			& \scriptsize\textbf{\texttt{nDTW}}~$\uparrow$
			& \scriptsize\textbf{\texttt{t-nDTW}}~$\uparrow$
			\\
			\midrule
			\scriptsize \texttt{1}
                & Ours~($d_n$=1)
                & 73.0
                & 62.1
                && 67.5
                 & 53.4
			\\
                \scriptsize \texttt{2}
			    & Ours~($d_n$=2)
                & 73.3
                & 62.6
                && 68.1
                 & 54.0
			\\
                \scriptsize \texttt{3}
			    & Ours~($d_n$=3)
                & 73.0
                & 62.3
                && 68.7
                 & 54.9
			\\
                \scriptsize \texttt{4}
			    & Ours~($d_n$=4)
                & 72.1
                & 61.2
                && 68.1
                 & 54.4 
			\\
                \scriptsize \texttt{5}
			    & Ours~($d_n$=5)
                & 71.9
                & 61.2
                && 68.0
                 & 54.2
			\\
			\bottomrule
		\end{tabular}}
	\caption{The performance of our method on IR2R with different neighbour distances $d_n$, which determines the size of neighbouring set. Our method achieves the best performance with $d_n = 3$.
	}
	\label{tab:depth_sensitivity}
\end{table}

\setlength{\tabcolsep}{.3em}
\begin{table}[t]
    \setlength{\aboverulesep}{0pt}
    \setlength{\belowrulesep}{0pt}
    \renewcommand{\arraystretch}{1.15}
    \centering
	\resizebox{0.8\linewidth}{!}{
		\begin{tabular}{cl cc c s c cs}
			\toprule
            & & &
			& \multicolumn{2}{c}{\scriptsize\textbf{Val-Seen}}
		   && \multicolumn{2}{c}{\scriptsize\textbf{Val-Unseen}}
            \\
			\cmidrule{5-6}
			\cmidrule{8-9}
			\scriptsize \shortstack{\#} &
			{\scriptsize Model}
                & \footnotesize \textsc{Distance}
			& \footnotesize \textsc{View Index}
			& \scriptsize\textbf{\texttt{nDTW}}~$\uparrow$
			& \scriptsize\textbf{\texttt{t-nDTW}}~$\uparrow$
			&
			& \scriptsize\textbf{\texttt{nDTW}}~$\uparrow$
			& \scriptsize\textbf{\texttt{t-nDTW}}~$\uparrow$
			\\
			\midrule
			\scriptsize \texttt{1}
                & Ours
                & &
                & 70.7
                & 58.6
                && 66.1
                 & 50.6
			\\
                \scriptsize \texttt{2}
			    & Ours
                & \checkmark &
                & 72.2
                & 61.3
                && 67.2
                 & 51.3
			\\
                \scriptsize \texttt{3}
			    & Ours
                & & \checkmark
                & 72.5
                & 62.0
                && 68.2
                 & 53.4
			\\
                \scriptsize \texttt{4}
			    & Ours
                & \checkmark & \checkmark
                & 73.0
                & 62.3
                && 68.7
                 & 54.9
			\\
			\bottomrule
		\end{tabular}
        }
	\caption{The ablation of omnigraph information. We remove the distance $d_v$ and viewindex $h_v$ for all keywords. Both distance $d_v$ and viewindex $h_v$ contribute to the performance improvement, which indicates the importance of positional information and structured memory.}
	\label{tab:structural_ablation}
\end{table}

Following previous studies~\cite{krantz2023iterative}, we present the experiment results on IR2R and IR2R-CE in this section.

\vspace{1mm}

\noindent \textbf{Dataset} IR2R~\cite{krantz2023iterative} is an iterative version of R2R~\cite{anderson2018vision} dataset, a common evaluation benchmark for Vision-and-Language Navigation in discrete environments collected from Matterport3D~\cite{chang2017matterport3d}. Same as R2R, IR2R is split into training/val-seen/val-unseen set. IR2R training set contains 14025 episodes from 61 scenes, and each scene has 3 tours. IR2R val-seen and val-unseen set has 53/11 scenes, 1011/2349 episodes, and 159/33 tours respectively. IR2R-CE~\cite{krantz2023iterative} is the iterative version of R2R-CE~\cite{krantz2020beyond} benchmark, which transforms the instruction and path in R2R to continuous form and uses Habitat~\cite{savva2019habitat} for environment simulation.

\vspace{1mm}

\noindent \textbf{Evaluation Metric} Following previous studies~\cite{krantz2023iterative}, we use t-nDTW~\cite{krantz2023iterative} to evaluate the performance of different methods in IVLN. t-nDTW is the tour-version of normalized dynamic time warping~(nDTW)~\cite{ilharco2019general}, which is a common metric for evaluation in VLN by measuring the normalized similarity between the ground truth path and the agent's navigation path. 
Please refer to more details in ~\cite{krantz2023iterative}.

\vspace{1mm}

\noindent \textbf{Implementation Details} We implement our method on ~\cite{krantz2023iterative} and retain all the hyper-parameters of VLN agents, including HAMT and MAP-CMA. The LLM we use for keyword extraction is GPT-3.5~\cite{brown2020language}, and the prompts are provided in supplementary materials. For OVD detection, we use OWL-ViT~\cite{minderer2022simple} with ViT-L/14 backbone~\cite{dosovitskiy2021image}. For discrete environments, we set neighbour distance $d_n$ as 3. For continuous environments, we set neighbour distance $d_n$ as 7, discovery threshold $d_{vp}$ as 1, and detection threshold $d_{det}$ as 0.25. 
All experiments are conducted on two RTX-3090 GPUs with 24GB memory. We train the model on one GPU and run the detection model on the other. Please refer to supplementary materials for details.

\vspace{1mm}

\noindent \textbf{Experiment Result} The experiment results on IR2R and IR2R-CE are shown in Table~\ref{tab:ivln_results} and Table~\ref{tab:ivlnce_results}. In Table~\ref{tab:ivln_results}, we compare our method to HAMT baseline and four different variants of TourHAMT. The performance of TourHAMT is copied from ~\cite{krantz2023iterative}. For each model, we present Total Length~(TL), Navigation Error~(NE), Oracle Success~(OS), nDTW, Success Rate~(SR), and Success weighted by Path Length~(SPL) besides t-nDTW. These six metrics are calculated on the episode level. t-nDTW is the tour-based metric for IVLN evaluation. As shown in Table~\ref{tab:ivln_results}, TourHAMT fails to improve the performance using tour navigation history, while our method gains 4\% and 5\% improvement in val-seen and val-unseen set respectively. Moreover, our method achieves better performance in episode-level metrics, including SPL, nDTW, SR, OS and NE. Note that our method is based on HAMT, the superior performance demonstrate the benefit of utilizing the history of previous episodes. 

Table~\ref{tab:ivlnce_results} shows the performance in continuous environments. We compare our method to CMA~\cite{krantz2020beyond} and its three variants, TourCMA, PoolCMA and PoolEndCMA~\cite{krantz2023iterative}. We also compare our method to MAP-CMA, a strong baseline for IVLN-CE. Our method gains 4\% and 3\% t-nDTW improvement in val-seen and val-unseen set respectively, while achieving better or comparable performance in episode-level metrics. 
Note that the performance of MAP-CMA~\cite{krantz2023iterative} is copied from the original paper. Our method is implemented on MAP-CMA with the same experiment setting, i.e., inferred semantics and iterative map construction, on which MAP-CMA achieves its best performance.

%% file: sec/5_ablation.tex
\section{Ablation Study}

\begin{figure*}
    \centering
    \begin{subfigure}[b]{0.49\textwidth}
        \includegraphics[width=\textwidth]{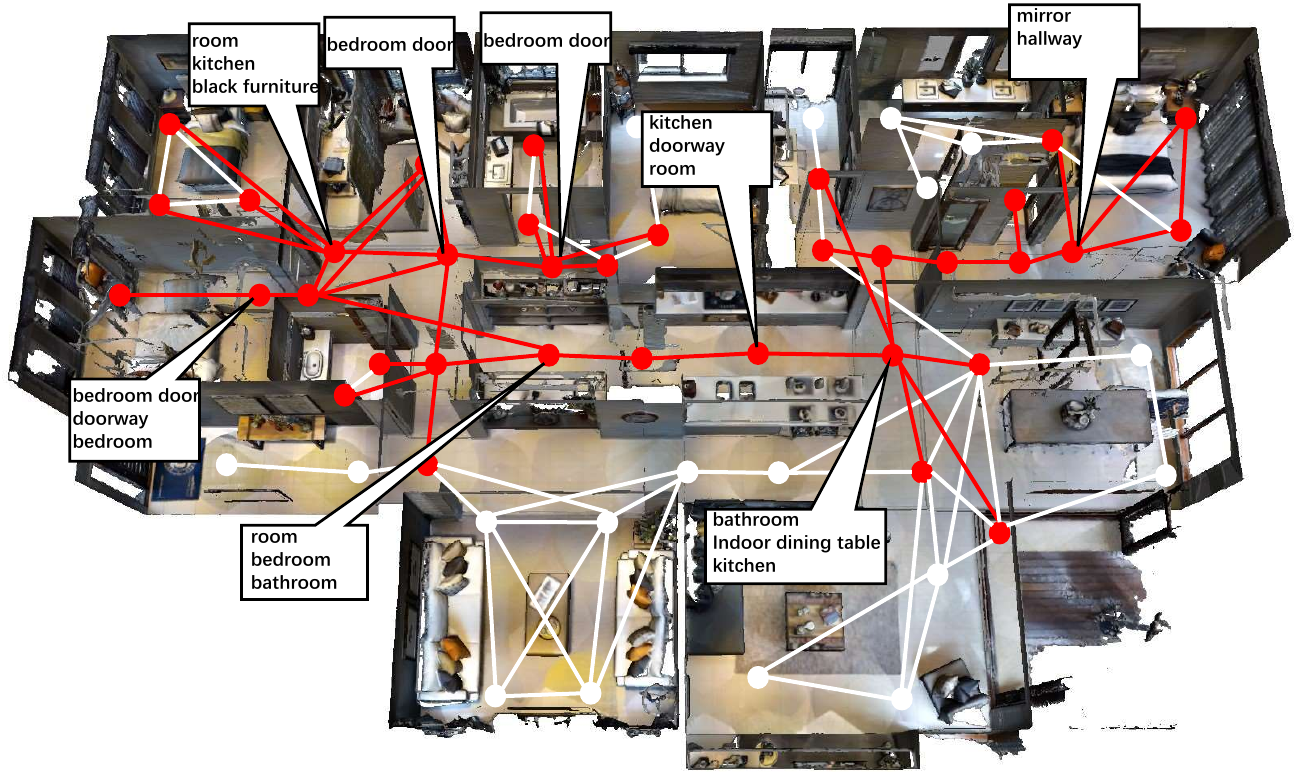}
        \caption{Omnigraph visualization after 5 episodes in a 100-episode tour.}
        \label{fig:omnigraph_5_episode}
    \end{subfigure}%
    ~ 
    \begin{subfigure}[b]{0.49\textwidth}
        \includegraphics[width=\textwidth]{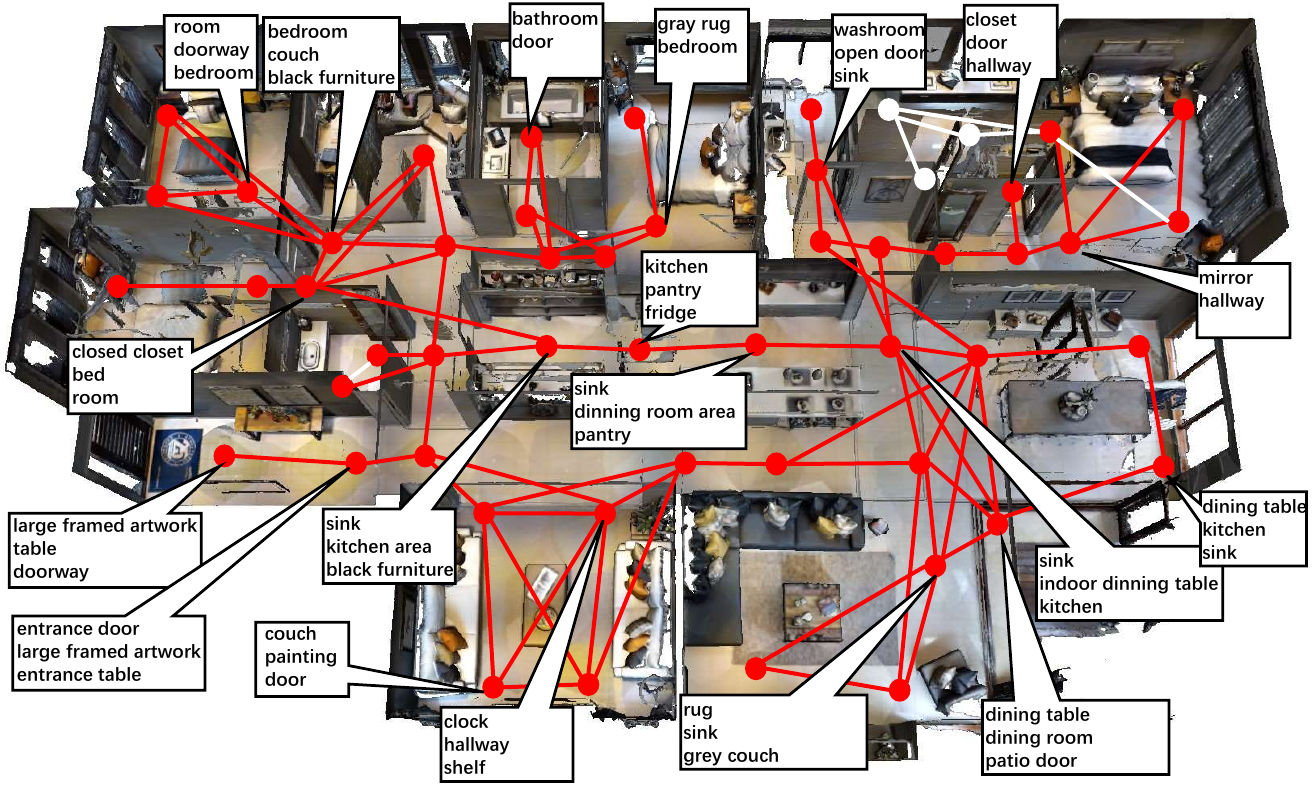}
        \caption{Omnigraph visualization after 50 episodes in a 100-episode tour.}
        \label{fig:omnigraph_50_episode}
    \end{subfigure}
    ~ 
    \caption{The visualization of Omnigraph in our method during a 100-episode tour. As the tour proceeds, the omnigraph becomes larger with more viewpoints and more connections. The keywords attached to viewpoints become more precise and diverse. We show 3 keywords for each viewpoint at most and omit the extra information~(e.g., heading) for simplicity.}\label{fig:omnigraph_visualization}
\end{figure*}


\subsection{Ablation on Open-Vocabulary Keywords}

To demonstrate the superiority of the open-vocabulary keywords in our method, here we ablate the Open-Vocabulary strategy in the three variants of our method in Table~\ref{tab:open_voca_ablation}.

\noindent\textbf{Type-I} Following previous studies\cite{cartillier2021semantic}, we collect and focus on the 12 most common object categories in R2R environment: \textit{\{chair, table, cushion, cabinet, shelving, sink, dresser, plant, bed, sofa, counter, fireplace\}} (sorted in descending order by number of object instances). Type-I discards the Keyword Extraction stage and performs the Keyword-Panoramic Detection with the above 12 categories as keywords. In this case, the OVD detector in our method detects the objects in the observations in a close-vocabulary way, i.e., all detection boxes sent to the agent's memory are limited to the 12 categories above. 

\noindent\textbf{Type-II} Compared to the original method, Type-II only sends the detection boxes with labels that contain at least one of the 12 most common objects above. For example, \textit{``marble kitchen counter''} contains \textit{``counter''} while \textit{``kitchen island''} does not contain any class names above. Thus only \textit{``marble kitchen counter''} will be sent to the detector and then to the agent's memory. In contrast, Type-I only has \textit{``counter''} in its keywords.

\noindent\textbf{Type-III} Type-III shows the performance of models with oracle keyword detections. After training the agent, we evaluate the agent twice on the validation seen/unseen dataset and do not clear the agent's memory after the first evaluation. Thus the agent's memory contains the detection result of the first evaluation which shares the same keywords to the second evaluation. Type-III simulates the scenarios where all the keywords in the evaluation have been detected and saved to the memory, while the parameters of the agent are not updated. 

Our method outperforms Type-I and Type-II models on both Val-Seen and Val-Unseen datasets. The difference between the four models indicates the performance contribution of open-vocabulary detection. Note that Type-III achieves the best performance because its memory contains the exact keywords of every episode in advance. 

\subsection{Ablation on Structured Memory}

To demonstrate the effectiveness of our structured memory, we ablate the two important attributes for each keyword in omnigraph, distance $d_v$ and view index $h_v$. The experiments are conducted in the discrete environment, i.e., IR2R. 

Specifically, OVER-NAV fuses distance $d_v$ and view index $h_v$ with each keyword. $d_v$ indicates the distance between the agent's current position and the viewpoint where the keyword is detected, while $h_v$ represents the direction the agent should move towards to go to the viewpoint where the keyword is detected. 
Table~\ref{tab:structural_ablation} shows the ablation experiment results. It can be observed that both attributes, distance $d_v$ and view index $h_v$ contribute to the performance improvement. The ablation of each attribute will decrease the performance, and the view index $h_v$ is more important than distance $d_v$, because view index $h_v$ indicates which direction the agent should move towards more clearly. 

\subsection{Sensitivity for Hyper-Parameter Depth $d_n$}

Hyper-parameter depth $d_n$ is the distance threshold for neighbour identification as described in Section~\ref{sec:method}. Neighbour identification only collects the viewpoints whose distances to the agent's current position are less than threshold $d_n$, and sends the collected viewpoints to the following procedures. Here we analyze the sensitivity for $d_n$. The performance of our method with different $d_n$ on IR2R is shown in Table~\ref{tab:depth_sensitivity}. The best performance is achieved when $d_n = 3$.

\subsection{Visualization}

We visualize the omnigraph in Fig.~\ref{fig:omnigraph_visualization} on IR2R. The tour shown in Fig.~\ref{fig:omnigraph_visualization} contains 100 episodes in the same scene, and Fig.~\ref{fig:omnigraph_5_episode} and Fig.~\ref{fig:omnigraph_50_episode} shows the omnigraph after 5 episodes and 50 episodes respectively. The dots/lines indicate the viewpoint and connectivity between them respectively. The red dots/lines are the viewpoints/connections visited by the agent in previous episodes, while the white dots/lines are unvisited. We show 3 keywords for some viewpoints at most and omit the extra information (e.g., heading) for simplicity.
As shown in Fig.~\ref{fig:omnigraph_5_episode}, the agent navigates a large portion of the scene~(66\% of all viewpoints) with a small number of episodes~(5\% of the instructions). 
The keywords from the 5 instructions are limited, but we can still observe that some open-vocabulary keywords are correctly detected, e.g., black furniture and indoor dining table. In Fig.~\ref{fig:omnigraph_50_episode}, most of the viewpoints are recorded in the omnigraph as well as the connectivity. The keywords attached to the viewpoints are more diverse and precise, which enhances the representability of the omnigraph and provides more accurate information for the agent.

%% file: sec/6_conclusion.tex
\section{Conclusion}

We propose an open-vocabulary-based method, OVER-NAV for Iterative Vision-Language Navigation. OVER-NAV incorporates LLMs and an Open-Vocabulary detector to construct an omnigraph, which consists of viewpoints and connections with keywords to describe the distribution of key objects that are important for the agent's navigation. Extensive experiments in both discrete and continuous environments demonstrate that omnigraph is a superior and more general structured memory to memorize and describe the navigation scene, enabling the agent to utilize the navigation history of previous episodes in IVLN for better performance. 

%% file: sec/X_suppl.tex
\clearpage
\setcounter{page}{1}
\maketitlesupplementary

\section{System Prompt for GPT in Keyword Extraction}

Here we provide the system prompt we use in keyword extraction in Table~\ref{tab:system_prompts}. After setting the system prompt, we send instructions to GPT and get responses containing the keywords separated by commas. Then we split the response and send the keywords for the following procedures.

\begin{table}[h]\centering
\begin{minipage}{1.0\columnwidth}
\centering
\begin{sectionbox}[]{System Prompt} 
    \centering
      \footnotesize
    \begin{tabular}{p{0.97\columnwidth} c}
You are a helpful assistant. You can help me by answering my questions. I will give you some instructions for vision-language navigation, you need to give me the key objects that are mentioned in this instruction. Key object is the noun or noun phrase that a navigation agent can use as milestone.

The query will be given by: 

Instruction: $\langle$QUERY$\rangle$

You must respond to any queries or answer in the following way:

Query: $\langle$QUERY$\rangle$ Answer: $\langle$ANSWER$\rangle$ Therefore the answer is: $\langle$TARGET\_OBJETCTS$\rangle$

The key objects in $\langle$TARGET\_OBJETCTS$\rangle$ must appear in the instruction and are separated by commas.
    \end{tabular}
\end{sectionbox}
\caption{System Prompt for keyword extraction from instructions.}
    \label{tab:system_prompts}
\end{minipage}
\end{table}

\section{Iterative REVERIE}

Here we further verify the effectiveness of our method on another navigation benchmark, REVERIE. 
REVERIE~\cite{qi2020reverie} is a benchmark for VLN with high-level instructions. The difference between REVERIE and R2R is that REVERIE replaces the instruction in R2R datasets with high-level instructions, which mainly describe the target location and objects. In contrast, R2R instructions provide detailed guidance to the agent along the ground truth navigation path. 

To evaluate the agent under the iterative vision-and-language navigation setting of REVERIE, the benchmark needs to be transformed into the \textit{iterative} version, which contains a tour file describing the episodes' order in which the instructions should be issued to VLN agents. 
Following~\cite{krantz2023iterative}, we generate the tours that minimize the distances between the end and starting points between the episodes in the tour. To this end, we employ Lin-Kernighan heuristic~(LKH)
, which is an efficient solver for the asymmetric travel salesman problem. 

The comparison between our method, OVER-NAV and the baseline, HAMT is shown in Table~\ref{tab:reverie_results}. Our method can still achieve better performance in the challenging setting. Note that we report the performance of the checkpoint with the highest \textbf{t-nDTW} scores for both models, which is different from the original paper of HAMT\cite{chen2021history}.

\section{Illustration of Two OVER-NAV Agents}

\subsection{OVER-NAV with HAMT}

\label{sec:ours_hamt_framework}

\begin{figure*}[t]
  \centering
   \includegraphics[width=0.8\linewidth]{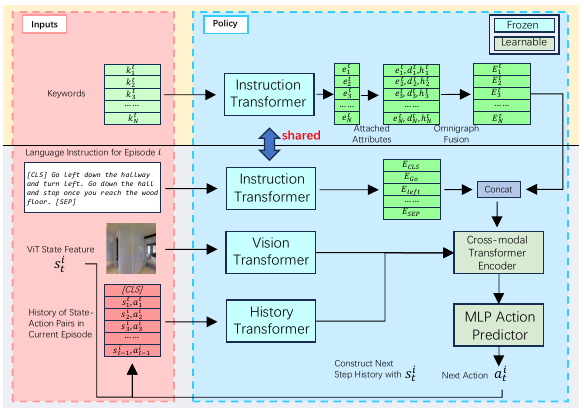}

   \caption{The overview of our method combined with HAMT.}
   \label{fig:overview_ours_hamt}
\end{figure*}

In Fig.~\ref{fig:overview_ours_hamt}, we present an overview of our method as applied to HAMT, the VLN agent for the discrete environment discussed in this paper. The illustration focuses on the data flow during step $t$ of episode $i$. The HAMT part of Fig.~\ref{fig:overview_ours_hamt} refers to IVLN\cite{krantz2023iterative}.

The upper and lower sections of Fig.~\ref{fig:overview_ours_hamt} depict OVER-NAV and HAMT, respectively. Within the HAMT framework, the language instruction for episode $i$ undergoes processing in the instruction transformer, generating an embedding sequence of equal length, inclusive of the [CLS] and [SEP] tokens. Each embedding in the sequence corresponds to the instruction word in the same position. Concurrently, the agent captures observations during navigation, forwarding them to the vision transformer to extract image features. 
The ViT state feature, denoted as $s_t^i$, is produced by the vision transformer using both observations and angles.
HAMT further maintains a history queue containing state-action pairs from previous steps within the current episode. 
The history transformer processes the history queue and generates the history embeddings.
HAMT employs a cross-modal transformer encoder to fuse cross-modal inputs, where instruction embeddings serve as the text modal, while history and observation embeddings function as the visual modal. Notably, the instructions transformer, vision transformer, and history transformer undergo pre-training on proxy tasks before being frozen during the navigation task training in HAMT.
Finally, the final model output is the action prediction $a_t^i$ for the current step. HAMT then appends the state-action pair of the current step, $s_t^i$ and $a_t^i$, to the history queue.


OVER-NAV prepares the keywords as described in Section~\ref{sec:method}. 
Subsequently, OVER-NAV leverages the same instruction transformer to derive embeddings for each keyword. Similar to the instructions, we add [CLS] and [SEP] tokens to each keyword, utilizing the embedding of the [CLS] token as the representation for the respective keyword. After the omnigraph fusion with the attached attributes, the keyword embeddings are arranged based on the distance metric $d_i^t$ and are appended to the instruction embeddings. The [SEP] token at the end of the instruction, serves as a separator between the two sections. Ultimately, the concatenated embedding functions as the text-modal representation and is transmitted to the cross-modal transformer encoder.

\subsection{OVER-NAV with MAP-CMA}

\begin{figure*}[t]
  \centering
   \includegraphics[width=0.8\linewidth]{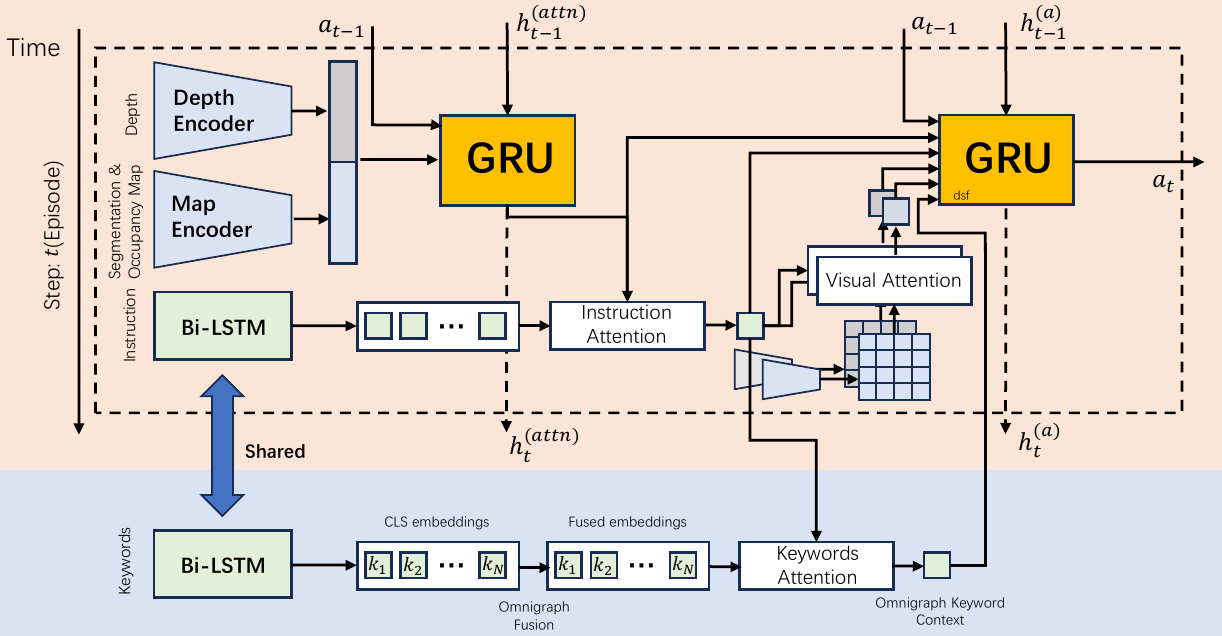}

   \caption{The overview of our method combined with MAP-CMA.}
   \label{fig:overview_ours_mapcma}
\end{figure*}

Fig.~\ref{fig:overview_ours_mapcma} illustrates the framework of our method when applied to MAP-CMA, the VLN agent in the continuous environment in this paper. The MAP-CMA part of Fig.~\ref{fig:overview_ours_mapcma} refers to IVLN~\cite{krantz2023iterative}.

The upper/bottom part of Fig.~\ref{fig:overview_ours_mapcma} shows MAP-CMA and OVER-NAV respectively. In MAP-CMA, the depth encoder encodes the depth images as depth embeddings and a bidirectional LSTM extracts the instruction embeddings from instructions. The segmentation map and occupancy map are concatenated and sent to the map encoder to produce the map embedding. The first GRU module serves as the state encoder, which encodes the depth embedding and map embedding at step $t$ as the state embedding. The instruction attention is performed with instruction embeddings and state embedding to text embedding. Later the text embedding is sent to the visual attention module, which performs attention on the feature maps of depth images and map images. The second state encoder, \ie, the GRU module, takes state embedding, text embedding, depth embedding, map embedding, and hidden state $h^{(a)}_{t-1}$ as inputs, and generates the predicted action $a_t$ and new hidden state $h^{(a)}_{t}$ as outputs. 

To incorporate OVER-NAV to MAP-CMA, we use the instruction bidirectional LSTM  for keyword embedding extraction. Each keyword is represented by the [CLS] token embedding. After omnigraph fusion, the positional information, \eg, heading and distance, is fused to the keyword embeddings. Then we perform the keywords attention with text embedding as the query. Finally, the omnigraph keyword context is sent to the second state encoder to aid the action prediction.
Similar to Section~\ref{sec:ours_hamt_framework}, the omnigraph keyword context provides the distribution information of detected objects in previous episodes.

\section{Code Implementation}

We further provide the code implementation of our method in discrete environments, \ie, on HAMT\cite{chen2021history} in the supplementary material.

\setlength{\tabcolsep}{.3em}
\begin{table*}[t]
    \renewcommand{\arraystretch}{1.15}
    \setlength{\aboverulesep}{0pt}
    \setlength{\belowrulesep}{0pt}
    \centering
	\resizebox{0.7\textwidth}{!}{
		\begin{tabular}{cl c ccccs c ccccs}
			\toprule
            & & 
			& \multicolumn{5}{c}{\scriptsize\textbf{Val-Seen}}
		   && \multicolumn{5}{c}{\scriptsize\textbf{Val-Unseen}}
            \\
			\cmidrule{4-8}
			\cmidrule{10-14}
			\scriptsize \shortstack{\#} &
			{\scriptsize Model}
			&
			& \scriptsize\textbf{\texttt{TL}}
			& \scriptsize\textbf{\texttt{OS}}~$\uparrow$
			& \scriptsize\textbf{\texttt{SR}}~$\uparrow$
			& \scriptsize\textbf{\texttt{SPL}}~$\uparrow$
			& \scriptsize\textbf{\texttt{t-nDTW}}~$\uparrow$
			&
			& \scriptsize\textbf{\texttt{TL}}
			& \scriptsize\textbf{\texttt{OS}}~$\uparrow$
			& \scriptsize\textbf{\texttt{SR}}~$\uparrow$
			& \scriptsize\textbf{\texttt{SPL}}~$\uparrow$
			& \scriptsize\textbf{\texttt{t-nDTW}}~$\uparrow$
			\\
			\midrule
			\scriptsize \texttt{1}
                & \texttt{HAMT}
                & 
                & 9.8 \scriptsize{$\pm$0.2}
                & 22 \scriptsize{$\pm$1}
                & 23 \scriptsize{$\pm$1}
                & 20 \scriptsize{$\pm$1}
                & 38 \scriptsize{$\pm$1}
                &
                & \hphantom{0}9.6 \scriptsize{$\pm$0.1}
                & 20 \scriptsize{$\pm$2}
                & 22 \scriptsize{$\pm$1}
                & 19 \scriptsize{$\pm$0}
                & 28 \scriptsize{$\pm$1}
			\\
                \midrule
                \scriptsize \texttt{6}
                & \texttt{Ours}
                & 
                & 9.8 \scriptsize{$\pm$0.3}
                & \bf 37 \scriptsize{$\pm$2}
                & \bf 40 \scriptsize{$\pm$1}
                & \bf 35 \scriptsize{$\pm$2}
                & \bf 44 \scriptsize{$\pm$1}
                &
                & \hphantom{0}8.9 \scriptsize{$\pm$0.2}
                & \bf 24 \scriptsize{$\pm$1}
                & \bf 25 \scriptsize{$\pm$2}
                & \bf 22 \scriptsize{$\pm$1}
                & \bf 30 \scriptsize{$\pm$1}
			\\
			\bottomrule
		\end{tabular}}
	\caption{
	    The comparison between HAMT and ours on REVERIE dataset.
	}
	\label{tab:reverie_results}
\end{table*}

